\documentclass[conference]{IEEEtran}
\IEEEoverridecommandlockouts
\usepackage{cite}
\usepackage{amsmath,amssymb,amsfonts}
\usepackage{algorithmic}
\usepackage{graphicx}
\usepackage{textcomp}
\usepackage{xcolor}
\usepackage{subfig}
\usepackage{pifont}
\usepackage{amssymb}
\usepackage{url}

\definecolor{lightgray}{gray}{0.8}
\def\BibTeX{{\rm B\kern-.05em{\sc i\kern-.025em b}\kern-.08em
    T\kern-.1667em\lower.7ex\hbox{E}\kern-.125emX}}
\begin{document}
% \pagenumbering{⟨arabic⟩}

\title{Cloud Is Closer Than It Appears: Revisiting the Tradeoffs of Distributed Real-Time Inference
}

\author{
    \IEEEauthorblockN{Pragya Sharma}
    \IEEEauthorblockA{\textit{pragyasharma@ucla.edu}} 
    \IEEEauthorblockA{\textit{University of California Los Angeles}}
    \and
    \IEEEauthorblockN{Hang Qiu}
    \IEEEauthorblockA{\textit{hangq@ucr.edu}}
    \IEEEauthorblockA{\textit{University of California Riverside}}
    \and
    \IEEEauthorblockN{Mani Srivastava$^\dag$ \thanks{$\dag $Author holds concurrent appointments as a Professor of ECE and CS (joint) at UCLA, and as an Amazon Scholar. This paper describes work performed at the UCLA, and is not associated with Amazon.}}
    \IEEEauthorblockA{\textit{mbs@ucla.edu}}
    \IEEEauthorblockA{\textit{University of California Los Angeles}}
}
\maketitle
\begin{abstract}
The increasing deployment of deep neural networks (DNNs) in cyber-physical systems (CPS) enhances perception fidelity, but imposes substantial computational demands on execution platforms, posing challenges to real-time control deadlines. Traditional distributed CPS architectures typically favor on-device inference to avoid network variability and contention-induced delays on remote platforms. However, this design choice places significant energy and computational demands on the local hardware. In this work, we revisit the assumption that cloud-based inference is intrinsically unsuitable for latency-sensitive control tasks. We demonstrate that, when provisioned with high-throughput compute resources, cloud platforms can effectively amortize network and queueing delays, enabling them to match or surpass on-device performance for real-time decision-making. Specifically, we develop a formal analytical model that characterizes distributed inference latency as a function of the sensing frequency, platform throughput, network delay, and task-specific safety constraints. We instantiate this model in the context of emergency braking for autonomous driving and validate it through extensive simulations using real-time vehicular dynamics. Our empirical results identify concrete conditions under which cloud-based inference adheres to safety margins more reliably than its on-device counterpart. These findings challenge prevailing design strategies and suggest that the cloud is not merely a feasible option, but often the preferred inference location for distributed CPS architectures. In this light, the cloud is not as distant as traditionally perceived; in fact, it is closer than it appears.

% In addition, our analytical and empirical evaluations also integrate environmental parameters such as vehicle speeds, obstacle types, and workloads to assess the feasibility and safety of deployments under dynamic operating conditions.

% By coupling low-power on-device controllers with cloud-hosted perception, systems can achieve energy-efficient, context-aware, and deadline-compliant control in safety- and time-critical scenarios. 
\end{abstract}

\vspace{-1em}
\section{Introduction}

The proliferation of intelligent cyberphysical systems (CPS), ranging from autonomous vehicles to smart surveillance, has placed unprecedented demands on real-time perception and control \cite{kim2017review}. Central to these systems is the integration of deep neural networks (DNNs) within the perception stack, enabling data-driven decision-making under environmental uncertainty. These models process high-dimensional and often noisy sensor input, generating semantically rich representations that, in turn, inform actuation strategies. However, this integration introduces substantial computational latency and variability, which pose challenges to the strict timing guarantees mandated by real-time CPS operations.

In contrast to traditional signal processing pipelines characterized by deterministic and predictable execution, DNN inference exhibits highly operating context-dependent latency profiles. Inference time can span tens to hundreds of milliseconds, depending on model architecture and execution platform, and such temporal uncertainty jeopardizes system responsiveness. This problem is exacerbated by the increasing sophistication of sensing techniques, such as multimodal fusion \cite{lahat2015multimodal}, which drive the need for more computationally intensive models. Although these advances improve perceptual accuracy, they also strain on-device computational resources, which are often constrained in terms of memory, processing throughput, and energy budget. Consequently, deploying large, high-performing models directly on end devices becomes infeasible for latency-sensitive CPS workloads.

To address these constraints, the community has shifted toward distributed CPS architectures that decouple sensing from inference. In this paradigm, sensor data are transmitted over the network to external compute nodes such as cloud datacenters, capable of executing complex models with higher accuracy and lower inference latency. This decoupling permits the use of state-of-the-art architectures without being bounded by the device's compute resources. However, this approach introduces new sources of delay stemming from network variability and resource contention on the remote server. 

Deployment strategies in distributed CPS generally follow one of two canonical approaches. The first, common in latency-critical applications, such as autonomous driving, requires that inference be performed locally on the device \cite{liu2019edge}. In this model, cloud offloading is considered only under exceptional circumstances, for example, when the task complexity exceeds the local processing capacity. Systems such as Waymo \cite{WAYMO} and Tesla’s FSD software \cite{talpes2020compute} are prime examples of this design philosophy, which favors tightly bounded and deterministic compute pipelines on onboard GPUs. However, while this approach ensures real-time responsiveness, it imposes considerable energy overhead: Inference workloads can account for a substantial share of total power consumption, which requires recharging every 4 to 6 hours \cite{gawron2018life, sudhakar2022data}.

The second paradigm is more prevalent in large-scale, non-real-time applications like video surveillance, where inference is performed in the cloud by default. For example, systems such as Amazon Rekognition \cite{Amazon-Reco} typically offload video data to centralized servers for processing. This model assumes that the application can tolerate long and variable inference delays, which excludes its use in latency-critical control loops.

However, this long-standing perception of cloud-based inference as incompatible with real-time applications, primarily due to concerns over high and unpredictable network latency, is becoming increasingly obsolete in light of recent technological advances. Modern GPUs offer dramatically lower inference times and can accommodate large-scale, high-accuracy models with consistent throughput. At the same time, advances in networking infrastructure, including 6G and local cloud zones \cite{aws-local}, have reduced round-trip latencies to the low tens of milliseconds \cite{charyyev2020latency}. Moreover, cloud datacenters benefit from rapid hardware refresh cycles and software stack updates, enabling quicker adoption of emerging accelerator architectures and inference optimizations than is feasible on already-deployed IoT or vehicular platforms.

In this paper, we challenge the prevailing assumption that on-device inference is categorically superior for latency-sensitive tasks in distributed CPS. We focus on safety-critical applications with stringent real-time requirements and pose a fundamental question: \textit{Can cloud-based inference, despite incurring network latency, match or even exceed the responsiveness of on-device computation in real-time control loops?}
To answer this, we develop a formal analytical framework and validate it through high-fidelity hardware-in-the-loop simulations. Our findings demonstrate that, under a range of deployment conditions, cloud-hosted inference can outperform on-device processing. This reveals the importance of holistically analyzing system delays along with operating context, thus motivating a rethinking of current design strategies. 

Specifically, we make the following contributions. 
\begin{enumerate}
    \item We develop a generalized analytical model for distributed inference in real-time perception-driven control loops. Our modeling captures the interplay between sensing frequency, inference delay, network latency, and system load, enabling a principled evaluation of the suitability of various combination of model and platform in deployment configurations.
    \item We implement an emergency braking application using the CARLA simulator to evaluate the relationship between response latency and application performance under realistic operating conditions. Our system includes real-time detection, cloud-hosted inference, and local actuation. This setup enables controlled experiments across diverse vehicle and obstacle dynamics, grounding theoretical claims in realistic, safety-critical scenarios.
    \item We demonstrate analytically and empirically that, across a range of workload and context conditions, cloud-based inference not only satisfies real-time control constraints but often outperforms on-board processing. This counterintuitive result highlights the need to rethink inference placement in latency-critical CPS.
\end{enumerate}
% Our findings challenge prevailing design strategies and advocate for energy-efficient embedded devices on the vehicle, in conjunction with cloud collaboration, as a viable, and at times, superior deployment strategy.

\section{Related Work}

% Cloud computing \cite{qian2009cloud} has gained steady popularity in the last two decades due to its ability to host large and complex models, as well as deliver cost-effective scalability to a range of applications. 
\textbf{Inference Placement:} The problem of determining the most appropriate deployment location for task-specific inference has generally been studied as a service placement problem. This problem covers deciding \textit{where} to run the inference \cite{huang2012dynamic}, \textit{when} to offload a task to a remote server \cite{barbera2013offload}, and \textit{how} to offload data efficiently\cite{jiang2021joint}. The service placement problem itself has been addressed through various optimization objectives: latency minimization\cite{latren2019collaborative}, energy efficiency\cite{ahvar2019estimating}, QoS \cite{he2013cost}, and accuracy guarantees\cite{li2021appealnet} or a combination thereof \cite{li2018jalad}. For example, Kang et al. \cite{kang2017neurosurgeon} explore accuracy-latency trade-offs through model partitioning and early exit strategies. In addition, techniques such as model compression for faster inference \cite{yang2022cnnpc} and parallelization for higher throughput \cite{warneke2009nephele} have also emerged. Despite this breadth of research, placement strategies for latency- and safety-critical applications have traditionally favored on-device execution due to concerns about network delays and cloud reliability \cite{whaiduzzaman2014survey, ma2020exploring, sharma2023impact, sharma2025towards}. 

\textbf{Analytical Modeling:} Several studies have employed analytical models to capture system-level trade-offs in inference latency and reliability \cite{bruneo2013stochastic, gandhi2010optimality, gandhi2013exact, khazaei2011performance}. Salem et al. \cite{salem2023toward} propose a latency-accuracy optimization framework to allocate ML models across distributed nodes, while Varma et al. \cite{varma2012performance} and Ali-Eldin et al. \cite{ali2021hidden} apply queuing-theoretic approaches to establish feasibility bounds for cloud-based inference. Notably, Ali-Eldin et al. demonstrate that, under moderate workloads, cloud platforms can outperform local deployments due to superior compute parallelism and reduced queuing. However, work that captures the dynamic operating context of the application remains limited, especially for safety-critical scenarios. 

% For instance, Sun et al. \cite{sun2019adaptive} only offloads complex tasks to the cloud but performs most tasks locally. 

% Some research challenges this stance by favoring cloud but relying on local as a fallback controller. Fang et al. (IEEE TMC 2021) show that with predictable network conditions and powerful back-end servers, cloud inference can outperform on-device computation in both latency and accuracy. Complementary to this, Zhao et al. (MobiSys 2022) introduce a dynamic offloading framework that selects between cloud and device based on run-time inference delay predictions and workload statistics.

Moreover, many prior models assume the cloud as a static, high-latency resource. However, cloud computing continues to evolve rapidly, with leading providers, such as AWS \cite{aws} and Google Cloud \cite{google}, now offering advanced GPU instances capable of high-throughput inference. Furthermore, data centers are increasingly deployed at regional levels to bring compute closer to end-users \cite{aws-local}, aided by improved load balancing \cite{aws-load} and batching strategies \cite{aws-batch} that support multi-tenant inference workloads. These advancements allow the cloud to host and serve larger, more accurate models that generally better with unseen inputs \cite{jeong2020ood} and have faster refresh cycles. Currently, the network infrastructure is improving with the development of new technologies such as 6G \cite{du2020machine}, reducing transmission latencies and jitter. Taken together, these trends warrant a reevaluation of service placement decisions under a performance-centric lens. Our work develops an analytical framework to rigorously evaluate the viability of cloud inference for emergency response scenarios.

% Add analytical models, previous work

\section{System Model}

\begin{figure}[t]
    \setlength{\abovecaptionskip}{0cm}
    \centerline{\includegraphics[scale=0.25]{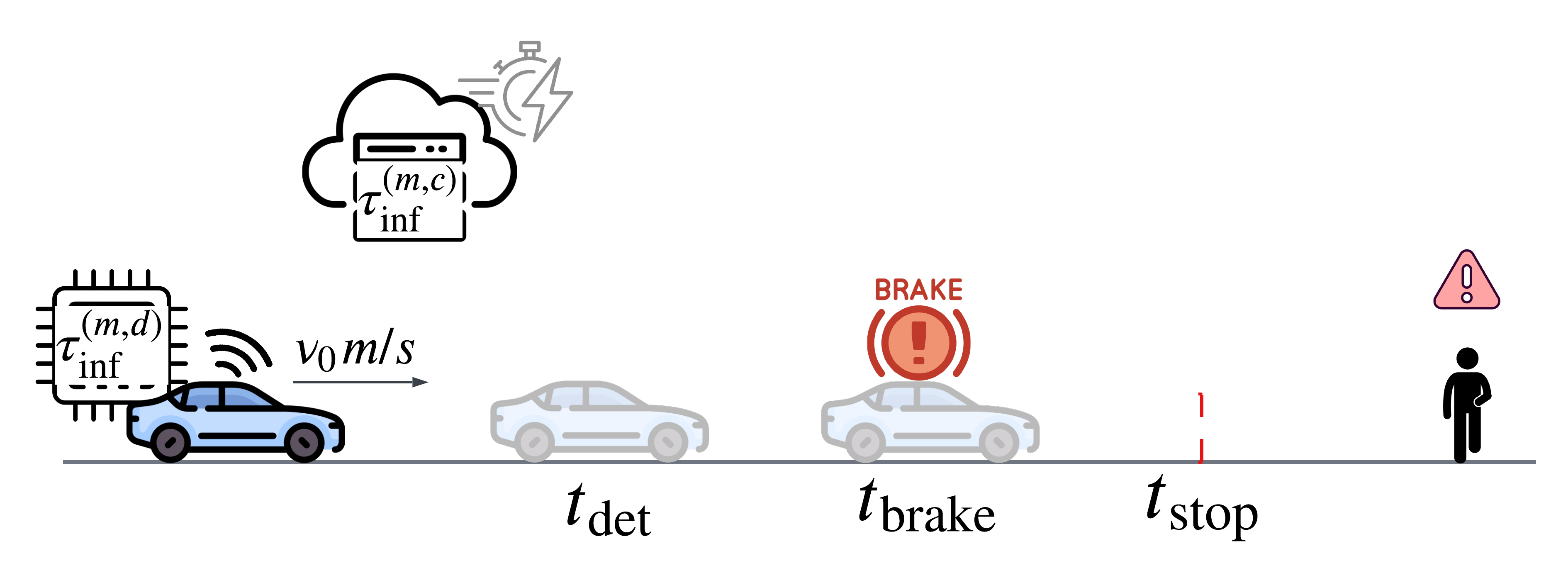}}
    \centering
    \caption{\small Temporal dynamics of emergency braking scenario. The ego vehicle traveling at $v_{0}$ m/s detects an obstacle at time $t_{\text{det}}$. A braking command is issued at $t_{\text{brake}}$ following inference execution either on-device $(m,d)$ or on cloud $(m,c)$ platform. The vehicle stops at $t_{\text{stop}}$.}
    \label{fig:sys}
    \vspace{-1.5em}
\end{figure}

\subsection{System Overview}

We consider a class of closed-loop, real-time CPS scenarios in which inferences derived from sensor data directly drive time-sensitive control actions. As a representative example, we focus on a safety-critical emergency braking task (Figure~\ref{fig:sys}), where an ego vehicle must detect and respond to stationary obstacles in its path. The perception pipeline is decoupled from actuation and can be executed on one of two processing platforms: on-device or cloud. Each captured image frame is forwarded to one of these platforms, where an object detection model is utilized to determine whether an obstacle exists in the path of the vehicle. If a potential obstacle is consistently detected across multiple frames, a braking signal is issued to the controller of the ego vehicle, which immediately initiates a braking maneuver. 

\subsection{Compute Fabric}
%Add info about types of machines used here and typical delays. 
The emergency braking system is supported by a two-tier computational architecture, consisting of on-device and cloud platforms. Each platform differs in terms of location, processing capacity, network delay, and energy constraints.

The \textit{on-device platform} is physically integrated within the vehicle and is positioned in conjunction with the sensors and actuators. It offers immediate access to sensor data, thereby eliminating network latency. However, it is limited in computational capacity and operates under tight energy budgets, which restrict the size and complexity of models it can run efficiently.

The \textit{cloud platform}, on the other hand, refers to remote datacenters equipped with powerful high-end GPUs. This tier supports the largest and most accurate models with fast inference times but introduces network delays due to its location. 

\subsection{Deployment-Time Feasibility}\label{feasibility}

We model the braking system as a real-time perception-to-action pipeline operating under a periodic sensing scheme. The camera sensor captures image frames at a fixed rate of \( F \) frames per second, producing an interval between frames \( \Delta = \frac{1}{F} \). This interval serves as a deployment time deadline $T_{max}$, used to assess whether candidate inference configurations are capable of responding within the available time between successive frames.

Let \( \mathcal{M} \) denote the set of available object detection models, and let \( x \in \{d, c\} \) denote the compute platform - on-device or in the cloud. Each model \( m \in \mathcal{M} \) can be deployed on one or more platforms, forming a set of model-platform pairs, \[ (m, x) \in \mathcal{M} \times \{d, c\} \] Each pair is characterized by a set of performance metrics such as the inference delay \( \tau_{\text{inf}}^{(m,x)} \), the energy consumed per inference \( E^{(m,x)} \), and the detection accuracy \( a^{(m,x)} \).

The total response latency for a pair \( (m, x) \) is given by
\begin{equation}
    T_{m,x} = \tau_{\text{nw}}^{(x)} + \tau_{\text{inf}}^{(m,x)} + \tau_{\text{ctrl}}
\tag{1}\label{time}
\end{equation}
where \( \tau_{\text{nw}}^{(x)} \) is the round-trip network delay for platform \( x \), and \( \tau_{\text{ctrl}} \) is the platform-agnostic control actuation delay. For the deployment on the device (\( x = d \)), we assume \( \tau_{\text{nw}}^{(d)} = 0 \). Although additional delays, such as sensor capture or sensor-to-application delay, contribute to the total response time, they are excluded from this formulation, as they remain invariant across platforms and do not affect the relative comparison.
To ensure real-time compliance under average conditions, we define the \textit{deployment-time feasibility set}:
\begin{equation}
    \mathcal{C} = \left\{ (m, x) \in \mathcal{M} \times \{d, c\} \;\middle|\; T_{m,x} \leq \Delta \;\land\; E^{(m,x)} \leq E_{\text{max}}^{(x)} \right\} \tag{2}
\end{equation}
where \( E_{\text{max}}^{(x)} \) is a platform-specific energy budget. This constraint is enforced for energy-constrained on-device platforms but may be relaxed for cloud deployments.

In addition, we select the optimal configuration by maximizing the detection accuracy in the feasible set such that:
\[
(m^*, x^*) = \arg\max_{(m,x) \in \mathcal{C}} a^{(m,x)}
\tag{3}\]

This deployment-time selection policy ensures that the chosen model-platform pair can meet both latency and energy constraints while prioritizing the highest achievable inference accuracy. However, at run-time, inference executes asynchronously. Once a frame is dispatched, it is processed to completion regardless of whether the result returns within the inter-frame deadline. This model reflects realistic execution behavior, allowing for variability in network conditions, resource contention, and queuing while preserving static feasibility guarantees established at deployment.

\section{Analytical Modeling}\label{analytical-model}
The feasibility set described in Section \ref{feasibility} enables model-platform selection under average-case assumptions. That is the latency, energy, and accuracy metrics used to define feasibility are typically profiled in isolation or under mean conditions. While such an approach may be sufficient for systems operating in stable environments, it fails to capture the variability inherent in real-world deployments. An alternative is to design for worst-case latency to ensure real-time constraints are met under all conditions. However, this conservative strategy can result in significantly degraded performance for typical or best-case scenarios, as the system may be forced to deploy suboptimal models in the interest of safety margins.

To address this, we develop analytical models that characterize how variability in system conditions, such as network and platform-specific delays, impact the timing and effectiveness of the braking task in real-world deployments.

\textit{\textbf{Lemma 1}: Cloud inference yields lower response latency than device inference when the network transmission delay is bounded by the difference in queue-amortized inference latency between the two platforms.\\}
% In particular,
% \begin{equation}
% \tau_{\text{nw}}^{(c)} < \frac{\tau_{\text{inf}}^{(m,d)}}{1 - F \cdot \tau_{\text{inf}}^{(m,d)}} - \frac{\tau_{\text{inf}}^{(m,c)}}{1 - F \cdot \tau_{\text{inf}}^{(m,c)}}
% \tag{4}\label{lemma1}
% \end{equation}
\textbf{Proof}: Consider two deployment scenarios for a given model \(m \in \mathcal{M}\): one executing locally on the device \((m,d)\), and the other executing remotely on a cloud server \((m,c)\). Although the model \(m\) may differ across platforms in practice, we adopt a unified notation for simplicity and denote both instances with \(m\). This abstraction does not affect the validity of our results. As previously defined in Eq. \ref{time}, the total response latency for each configuration \((m,x)\) is given by,
\[
T_{m,x} = \tau_{\text{nw}}^{(x)} + \tau_{\text{inf}}^{(m,x)} + \tau_{\text{ctrl}}
\]
% where $t_{\text{nw}}^{(x)}$ is the round-trip network delay to platform $x$, $t_{\text{inf}}^{(m,x)}$ is the inference latency, and $t_{\text{ctrl}}$ is the fixed control actuation time.

Since actuation occurs locally, \(\tau_{\text{ctrl}}\) is invariant across platforms. Additionally, on-device execution incurs zero network overhead. Thus, the response latencies are simplified as:
% \vspace{-0.3cm}
\begin{align*}
T_{m,d} &= \tau_{\text{inf}}^{(m,d)} + \tau_{\text{ctrl}} \\
T_{m,c} &= \tau_{\text{nw}}^{(c)} + \tau_{\text{inf}}^{(m,c)} + \tau_{\text{ctrl}}
\end{align*}

Then, for the cloud response time to be faster, we have:
\begin{equation*}
\tau_{\text{nw}}^{(c)} < \tau_{\text{inf}}^{(m,d)} - \tau_{\text{inf}}^{(m,c)}
\end{equation*}

Additionally, to incorporate the effects of system load, we assume inference requests arrive at a constant rate \(F\). We model each compute platform with a simple M/M/1 queue \cite{vilaplana2014queuing} with service rate \(\mu_x = 1 / \tau_{\text{inf}}^{(m,x)}\), under the constraint \(F \cdot \tau_{\text{inf}}^{(m,x)} < 1\). The expected queue-amortized inference latency on platform \(x\) is then:
\vspace{-0.3cm}
\begin{equation*}
\tau_{\text{inf}}^{(m,x)} = \frac{\tau_{\text{inf}}^{(m,x)}}{1 - F \cdot \tau_{\text{inf}}^{(m,x)}}
\end{equation*}

Substituting into the inequality above yields:
\begin{equation}
\tau_{\text{nw}}^{(c)} < \frac{\tau_{\text{inf}}^{(m,d)}}{1 - F \cdot \tau_{\text{inf}}^{(m,d)}} - \frac{\tau_{\text{inf}}^{(m,c)}}{1 - F \cdot \tau_{\text{inf}}^{(m,c)}}
\tag{4}\label{lemma1}
\end{equation}

We explicitly include queuing delays for the on-device latency model, despite the platform being dedicated, because queuing can still occur when inference delay exceeds the frame inter-arrival interval. Although such cases should be precluded via deployment-time feasibility checks, runtime perturbations such as thermal throttling or transient contention, can violate baseline assumptions. Capturing this possibility is essential for modeling worst-case latency and ensuring robust system behavior under variable conditions.

\textbf{Discussion: }This lemma establishes a sufficient condition under which cloud execution \((m,c)\) yields lower total response latency compared to device execution \((m,d)\). Specifically, when the round-trip network latency is smaller than the discrepancy in effective (i.e., queue-amortized) inference latency between the two platforms, cloud inference outperforms local inference. This has two important implications. First, in practical scenarios where \(\tau_{\text{inf}}^{(m,c)} \ll \tau_{\text{inf}}^{(m,d)}\), the condition is easily satisfied even in the presence of moderate network delays. Second, as the frame rate \(F\) increases, queuing delays on the device dominate the total latency, causing the right-hand side of the inequality to grow rapidly. This, in turn, enlarges the set of network conditions under which cloud-based inference becomes the latency-optimal choice.
This result demonstrates that the cloud platform, despite being remote, can achieve lower response latency than local execution, provided that its service rate advantage adequately amortizes the network delay.

\textit{\textbf{Lemma 2}: Inference and network delays induce temporal misalignment in control actuation, which can violate safety constraints even when control latency is platform-invariant.}

\textbf{Proof}: Let \( v_0 \) denote the initial velocity of the vehicle, and let \( a \) denote the magnitude of deceleration achieved once the braking begins. Using a constant deceleration model, the total distance required to come to a complete stop is given by
\[
s_{\text{stop}} = v_0 \cdot (\tau_{\text{nw}}^{(x)} + \tau_{\text{inf}}^{(m,x)}) + \frac{v_0^2}{2a}
\]
where the first term represents the distance traveled by the vehicle during the processing delay, and the second term captures the braking distance once the deceleration is initiated.

Let \( s_{\text{avail}} \) denote the distance from the vehicle to the obstacle at the time of detection. The braking decision is successful in preventing a collision if and only if,
\[
s_{\text{stop}} < s_{\text{avail}}
\]

This inequality can be equivalently expressed as a bound on the maximum allowable perception delay:
\begin{equation}
\tau_{\text{nw}}^{(x)} + \tau_{\text{inf}}^{(m,x)} < \tau_{\text{react}}
\tag{5}\label{lemma2}
\end{equation}
where the reaction-time budget \( \tau_{\text{react}} \) is defined as
\[
\tau_{\text{react}} = \frac{s_{\text{avail}}}{v_0} - \frac{v_0}{2a}
\]

\textbf{Discussion:} This result reveals that inference and network delays, while not directly increasing control latency, shift the control decision point beyond the physical window for safe intervention. Importantly, $\tau_{\text{react}}$ is not fixed but depends on the running environment. In the braking application, the run-time context includes factors such as vehicle speed and braking capability, and uncertainty in obstacle detection. Thus, a model-platform pair that satisfies the feasibility of deployment time (that is, $T_{m,x} \leq \Delta$) may still lead to unsafe outcomes if $\tau_{\text{nw}}^{(x)} + \tau_{\text{inf}}^{(m,x)} \geq \tau_{\text{react}}$ under specific driving conditions. We therefore introduce this condition as a runtime safety constraint that augments the static feasibility criterion and exposes a hidden coupling between perception latency and actuation efficacy in real-time control loops.

\textit{\textbf{Lemma 3}: Detection time is model- and platform-dependent, and this variability further shifts the timing of control actuation.}

\textbf{Proof:} Lemma 2 assumes that all the model-platform configurations observe the obstacle at a common reference time, defined by the moment the object becomes physically visible in the scene. However, in practice, the time of first detection is not constant between configurations but depends on the accuracy, robustness, and sensitivity of the deployed model. We therefore define the detection time $t_{\text{det}}^{(m,x)}$ as the frame time at which configuration $(m,x)$ first produces a valid detection with sufficient confidence.

The time\footnote{We use $t$-based notation to denote specific time instants, and $\tau$-based to denote time durations or intervals.} at which the control action is ultimately triggered is then given by:
\[
t_{\text{brake}}^{(m,x)} = t_{\text{det}}^{(m,x)} + \tau_{\text{nw}}^{(x)} + \tau_{\text{inf}}^{(m,x)}
\]

% Compared to the formulation in Lemma 2, which assumes detection time is aligned across configurations, this expression introduces an additional temporal offset. A model-platform pair that detects the obstacle later—even if it infers quickly and incurs low network latency—may still result in delayed actuation.

To ensure safe braking, we require that the entire delay from the physical appearance of the obstacle to the control actuation remain within the reaction time budget. Let $t_{\text{obs}}$ denote the time at which the obstacle first enters the scene and becomes theoretically detectable. Then, 
\[
t_{\text{brake}}^{(m,x)} = t_{\text{obs}} +  \tau_{\text{det}}^{(m,x)} + \tau_{\text{nw}}^{(x)} + \tau_{\text{inf}}^{(m,x)}
\]

where $\tau_{\text{det}}^{(m,x)}$ = $t_{\text{det}}^{(m,x)} - t_{\text{obs}}$ denotes the detection delay. Including this delay in the reaction time budget, we get:
\begin{equation}
\tau_{\text{det}}^{(m,x)} + \tau_{\text{nw}}^{(x)} + \tau_{\text{inf}}^{(m,x)} < \tau_{\text{react}}
\tag{6}\label{lemma3}
\end{equation}

\textbf{Discussion:} This result highlights an important and often overlooked dynamic: total response latency includes not only model execution and communication delay, but also the system’s ability to recognize and react to visual stimuli in a timely fashion. Delayed detection reduces the effective decision window, even in the presence of fast inference or proximity to the actuator.

Thus, safety in perception-to-action systems must be evaluated not only with respect to platform latency but also with respect to model-specific detection delay, which may vary significantly across configurations due to model accuracy and detection thresholds. This extension further tightens the run-time feasibility envelope and motivates end-to-end evaluations that jointly consider detection timeliness and response latency.

Although $t_{\text{obs}}$ is generally not observable in real-world systems, the lemma remains practically useful in two key ways. First, relative detection times across model-platform pairs are observable and actionable. Even if the absolute moment of appearance of the obstacle is unknown, earlier detection by one configuration compared to another provides a measurable advantage in response time. Second, $t_{\text{obs}}$ can be approximated offline using labeled sequences or simulation. These approximations allow system designers to empirically bound detection uncertainty and incorporate detection delays into deployment-time or run-time feasibility assessments. As such, this lemma provides the foundation for context-aware safety analysis, even in the absence of ground-truth timing.
\section{Experimental Evaluation}

In this section, we empirically validate the analytical results derived in Section \ref{analytical-model}. Specifically, our objective is to quantify the conditions under which cloud-based inference outperforms on-device execution in latency-sensitive control tasks. 

\subsection{System Setup}

We conducted experiments using the CARLA simulator \cite{dosovitskiy2017carla}, a high-fidelity platform to model urban driving environments with realistic sensor suites and traffic dynamics. The simulated ego vehicle is equipped with a front-facing RGB camera operating at a fixed sampling rate of $10$ frames per second. To capture variations in vehicle dynamics and deceleration profiles, we evaluated three distinct vehicle types. Audi (car), Carla Cola (truck), and Kawasaki Ninja (motorbike). Each vehicle is evaluated at speeds of $20$, $40$, and $60~\mathrm{mph}$, corresponding to urban, city, and highway driving conditions.

\begin{figure}[t]
\vspace{-1em}
    \centering
    \subfloat{
        \includegraphics[width=0.49\columnwidth]{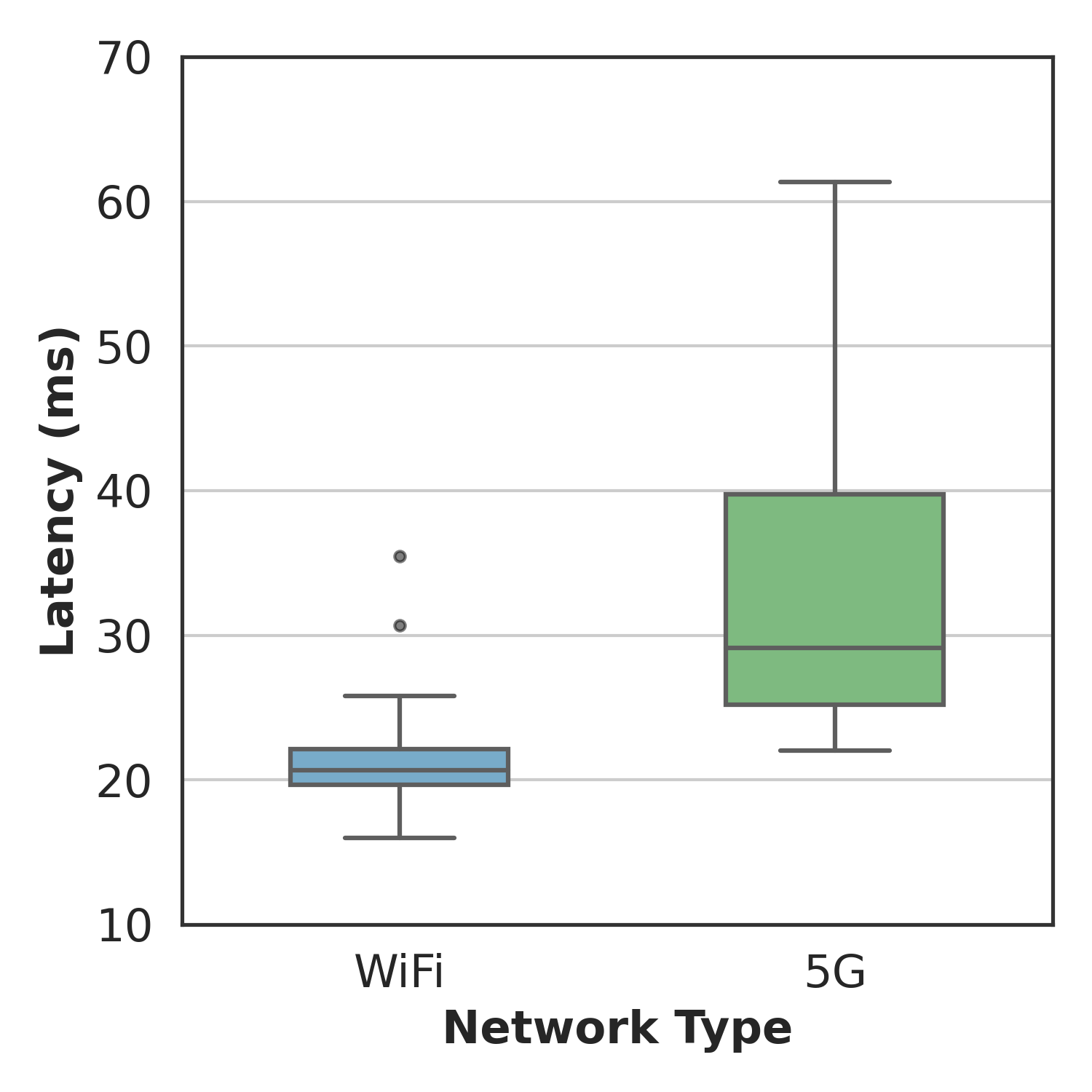}
        \label{fig:plot1}
    }
    % \hfill
    \subfloat{
        \includegraphics[width=0.49\columnwidth]{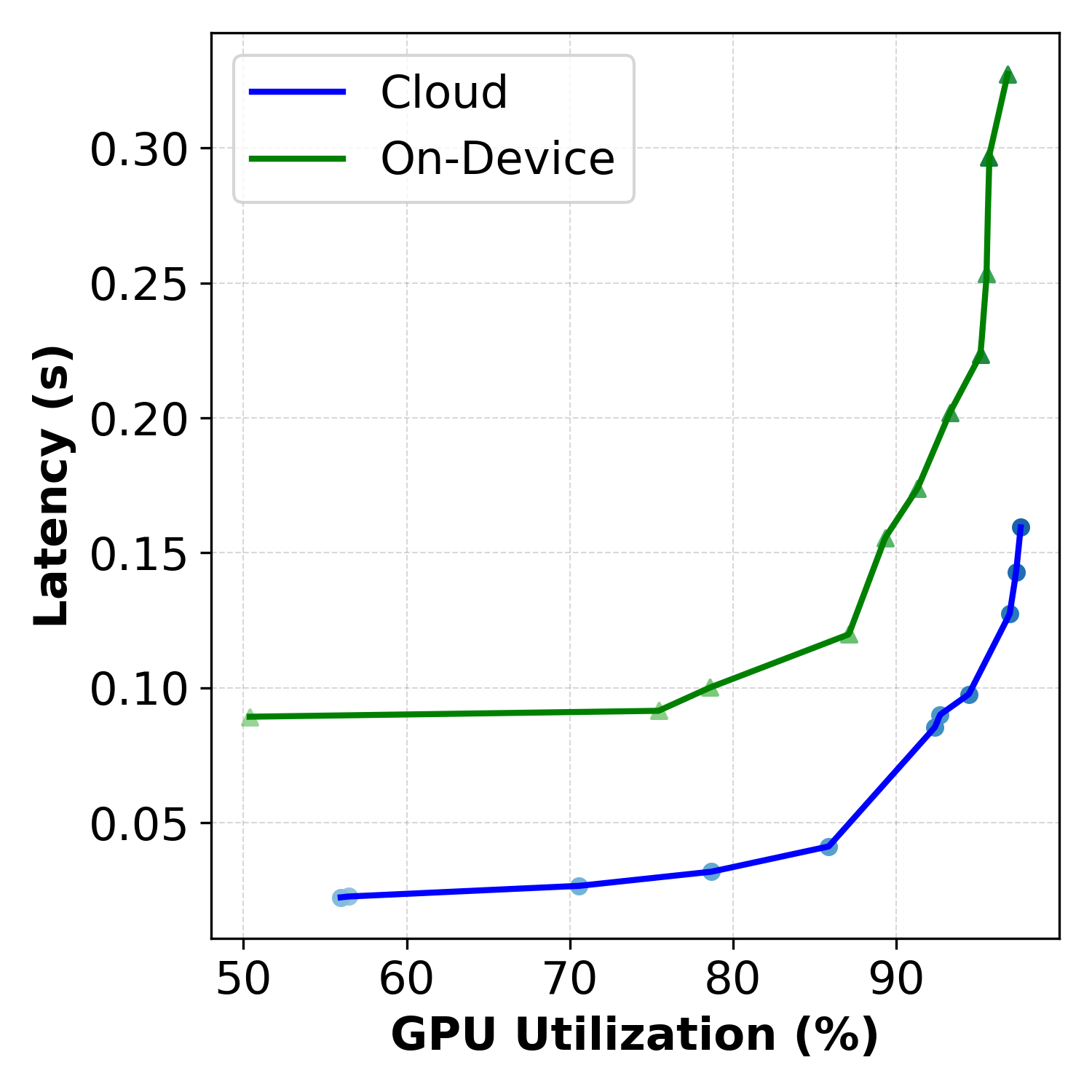}
        \label{fig:plot2}
    }
    \caption{\small(Left) Latency distributions for WiFi and 5G networks; (Right) Inference latency as a function of GPU utilization for cloud and on-device deployments.}
    \label{fig:standalone}
\vspace{-1.5em}
\end{figure}

To evaluate the perception-to-action latency in safety-critical settings, we introduce a static obstacle approximately $300~\mathrm{m}$ ahead of the vehicle’s trajectory. The perception module uses the YOLO11 \cite{yolo11_ultralytics} family of object detectors, which are pre-trained on the COCO dataset \cite{lin2014microsoft}. These models are evaluated in two deployment configurations: (1) on-device inference using an NVIDIA Jetson AGX Orin, and (2) cloud-based inference hosted on a dedicated RTX A5000 GPU server. In both configurations, the control logic remains local to ensure deterministic actuation latency.

Note that we do not deploy the perception stack directly on public cloud infrastructure due to the uncontrollable variability in latency and compute availability arising from time-of-day effects and background tenant load \cite{lu2017imbalance}. Therefore, to emulate realistic network-induced delays, we inject synthetic round-trip latencies that represent low ($p10$), median ($p50$), and high ($p90$) percentiles. These latency values are derived from empirical measurements obtained by transmitting CARLA-generated image frames ($20~\mathrm{kB}$ in size) to an instance of the AWS EC2 local zone located in the \texttt{us-west-2-lax-1a} region. The measurements include two network access technologies: urban Wi-Fi and 5G cellular networks. The summary statistics for each network class are visualized in Figure 2.

All system components operate within a controlled environment, with each vehicle initialized at a common starting location to ensure consistent measurement of critical events: frame capture\footnote{Frame capture marks the distance at which the image frame that eventually triggers detection was captured by the vehicle's camera.} (\scalebox{0.85}{\ding{72}}), brake execution (\scalebox{0.65}{\ding{108}}), and vehicle stop (\scalebox{0.65}{$\blacksquare$}). In addition, we record energy and throughput metrics to provide a comprehensive assessment of the trade-offs associated with model-platform selection for real-time CPS.

\subsection{Model Deployment}
To determine the optimal model-platform pairing for deployment, we empirically profile each variant in the YOLO11 model family using three key metrics: inference latency, detection accuracy, and energy consumption. The results, summarized in Table 1, are evaluated under a maximum target latency constraint ($T_{max}$) of $100~\mathrm{ms}$, chosen to align with a control loop time that remains safely below typical human reaction times, which range from 300 ms to 1.2 s \cite{johansson1971drivers}.

The detection accuracy (mAP) of each variant of the YOLO11 model remains consistent across deployment platforms, as it is derived from standardized benchmarks reported in the official YOLO11 documentation. Moreover, as expected, the on-device inference exhibits lower energy consumption across all variants, a consequence of the Jetson AGX Orin design, which prioritizes energy efficiency for resource-constrained environments. In contrast, the A5000 GPU is optimized for throughput, trading energy efficiency for raw computational power. This tradeoff is most evident in the inference latency: the A5000 achieves approximately $85\%$ lower inference times than the Jetson across all model sizes. Based on these results and guided by the feasibility conditions established in Section~\ref{feasibility}, we select YOLO11-medium for the deployment on the device and YOLO11-xlarge for the cloud deployment, since both configurations satisfy the constraint $T_{max}$.

\begin{table}[htbp]
\centering
\resizebox{\columnwidth}{!}{%
\begin{tabular}{c|c|cc|cc}
\hline
\textbf{Model} & \textbf{mAP} & \multicolumn{2}{c|}{\textbf{Cloud / A5000}} & \multicolumn{2}{c}{\textbf{On-Device / Jetson}} \\
\cline{3-6}
& & \textbf{Energy (J)} & \textbf{Latency (s)} & \textbf{Energy (J)} & \textbf{Latency (s)} \\
\hline
YOLO11x & 54.7 & 1.66 & 0.029 & 0.75 & 0.126 \\
YOLO11l & 53.4 & 1.27 & 0.028 & 0.75 & 0.126 \\
YOLO11m & 51.5 & 0.92 & 0.021 & 0.58 & 0.095 \\
YOLO11s & 47.0 & 0.76 & 0.019 & 0.43 & 0.088 \\
YOLO11n & 39.5 & 0.73 & 0.019 & 0.41 & 0.079 \\
\hline
\end{tabular}%
}
\label{tab:yolo11}
\end{table}
\vspace{0.5mm}
\noindent\parbox{\columnwidth}{\small{TABLE I: Comparison of YOLO11 models across platforms: accuracy, energy consumption, and inference latency}}

\subsection{Application Performance}
Figure 3(a) illustrates braking performance under baseline conditions for cloud and on-device deployments in a range of vehicle speeds and types. Although inference on devices eliminates network latency, it suffers from prolonged inference delays due to limited computational resources. In contrast, the cloud configuration, although it incurs a median round-trip delay of $22~\mathrm{ms}$, consistently initiates braking earlier and results in longer stopping distances from the obstacle. This superior performance stems from two key factors: (1) significantly lower inference latency enabled by the cloud's advanced computational infrastructure, and (2) the use of a larger model with higher sensitivity and accuracy. Cloud deployment detects obstacles almost $20~\mathrm{m}$ earlier by responding more effectively to subtle variations in input frames, reducing the delay between frame capture (\scalebox{0.85}{\ding{72}}) and the brake decision (\scalebox{0.65}{\ding{108}}). This gives the system more space to decelerate and respond safely within the available time budget $\tau_{\text{react}}$.
Such advantages are especially pronounced at higher speeds, where delays in perception and control compress available reaction time. As observed in the figure, the deployment of the device frequently results in stops near or within the unsafe braking zone at $40$ and $60~\mathrm{mph}$, reflecting the consequences of higher inference latency and reduced detection reliability. In contrast, cloud-based execution results in earlier brake commands even at higher speeds, promoting more consistent and safer outcomes.

\textbf{Takeaway:} These findings reinforce Lemmas 1-3 by empirically demonstrating that cloud-based inference, when operated under bounded and moderate network delay, can outperform on-device processing in both timeliness and safety for real-time cyberphysical systems.

\begin{figure*}[htbp]%
    % \setlength{\fboxrule}{0.1pt}
    % \begin{minipage}{0.75\textwidth}
    \begin{center}
    \fcolorbox{lightgray}{white}{\centering\includegraphics[width=0.9\textwidth]{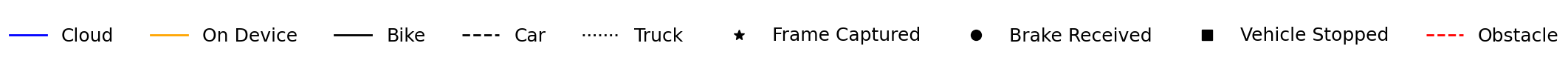}}
    \end{center}
    % \vspace{1mm}
    \subfloat[\centering \small Baseline scenario: Cloud enables early detection resulting in shorter braking and stopping distances than on-device inference.]
    {{\includegraphics[scale = 0.28]{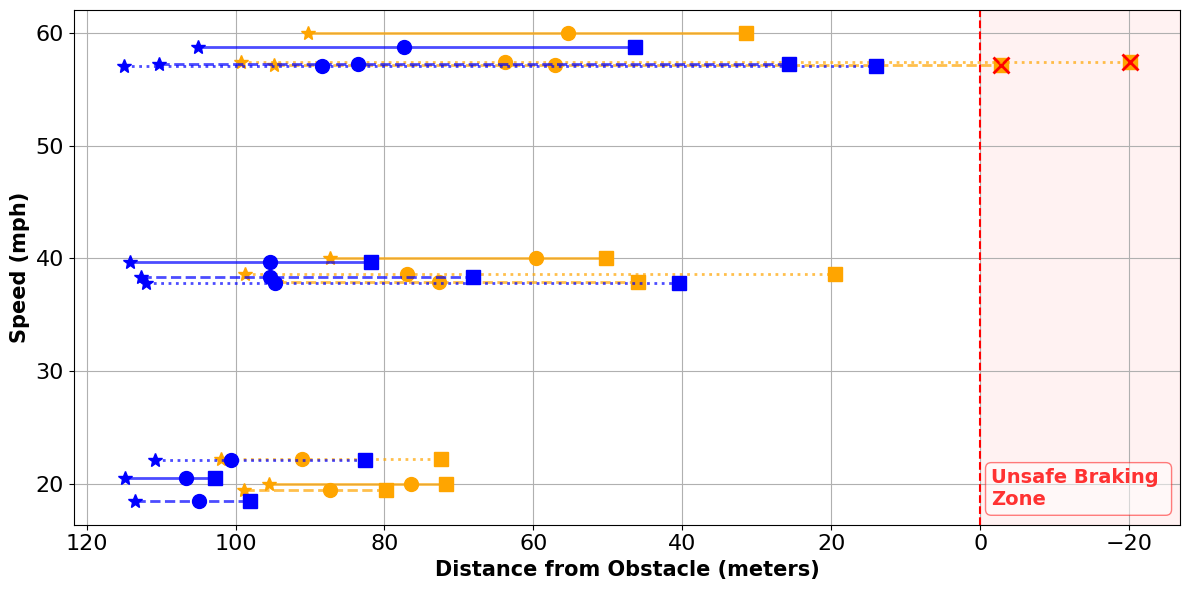} }}
    \label{figA}%
    \quad
    \subfloat[\centering \small Tail latency scenario: Cloud and on-device deployments exhibit comparable performance under high network latency.]
    {{\includegraphics[scale = 0.28]{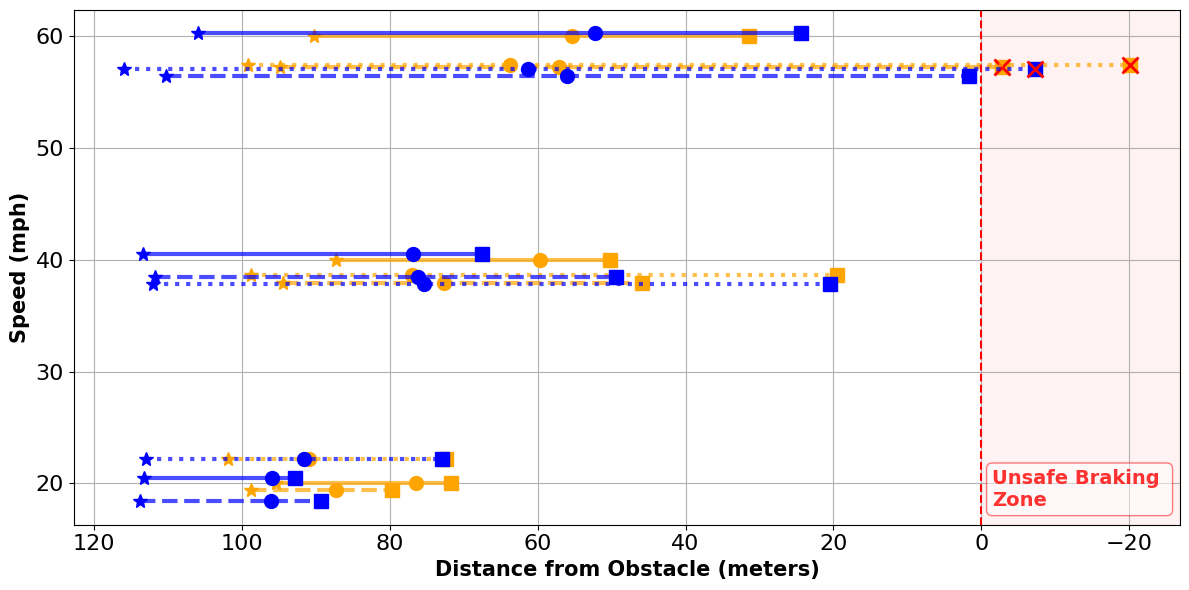} }}\label{figB}%
    \quad
    \subfloat[\centering \small Concurrent workload scenario: Cloud maintains stable braking performance under increased GPU contention.]{{\includegraphics[scale = 0.28]{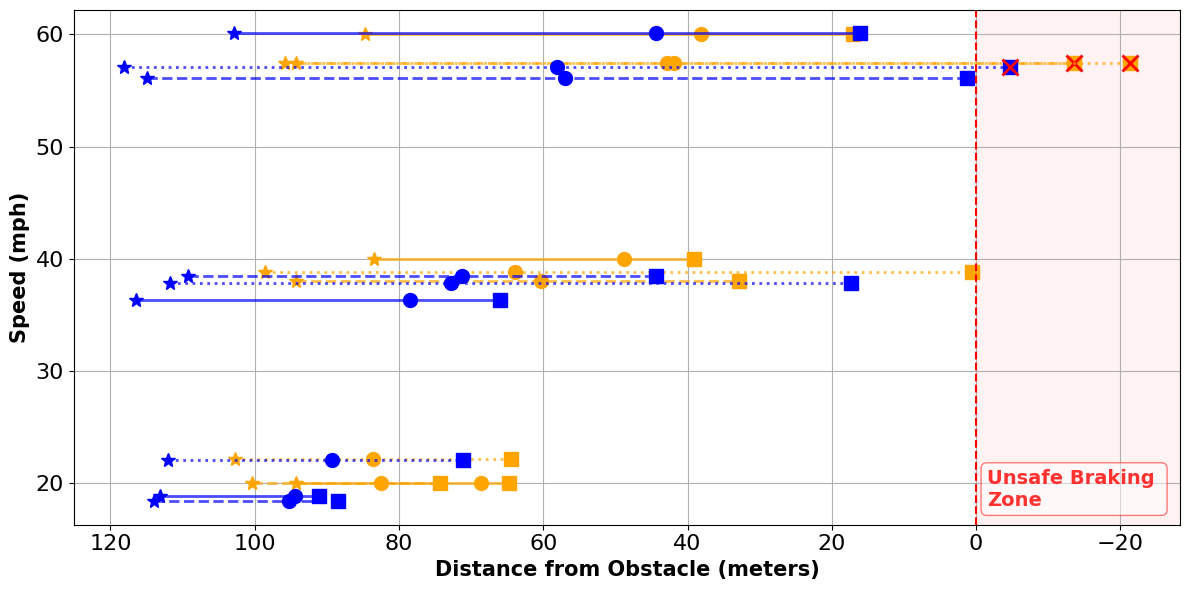} }}\label{figC}%
    \quad
    \subfloat[\centering \small Varying obstacle: Both deployments exhibit reduced detection ranges for smaller obstacles, narrowing the performance gap.]{{\includegraphics[scale = 0.28]{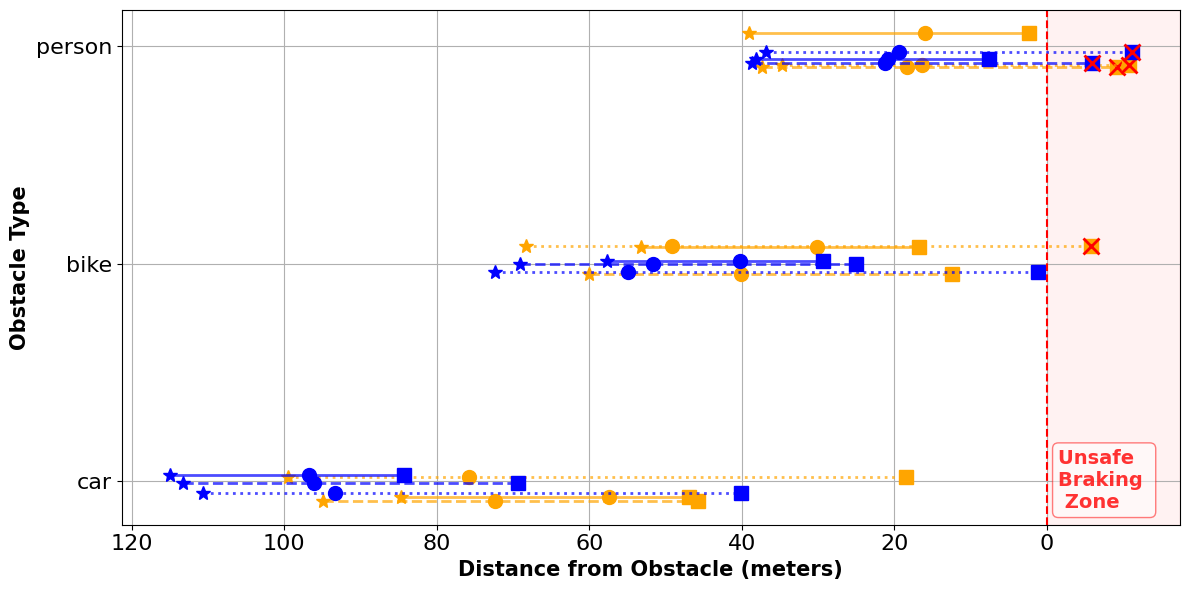} }}\label{figD}%
    \caption{\small Platform-wise braking performance under (a) baseline, (b) tail latency, (c) concurrent workload, and (d) varying obstacle scenarios. Each horizontal series corresponds to a specific vehicle–platform–speed configuration. Marker positions indicate distances at perception, brake reception, and vehicle stop. Line styles represent vehicle types, and unsafe outcomes are flagged in red.}%
    \label{fig:example}%
    %  \end{minipage}
    %  \vspace{-15mm}
    % \begin{minipage}{0.24\textwidth}
    %     \centering
    %     \includegraphics[width=\textwidth]{images/vertical_shared_legend.png}
    % \end{minipage}
    \vspace{-1em}
\end{figure*}

\subsection{Impact of High Network Latency}
To assess its robustness under adverse network conditions, we note the performance of the system when the cloud experiences tail network latency ($55-65~\mathrm{ms}$). Based on our measurements in Figure \ref{fig:standalone}, this represents rare but realistic conditions, such as network congestion or deployments geographically distant from data centers. In particular, the obstacle is observed for the first time at the same location (\scalebox{0.85}{\ding{72}}) in both baseline and adverse configurations. Only the timing of the actuation, specifically, the interval between frame capture (\scalebox{0.85}{\ding{72}}) and brake signal reception (\scalebox{0.65}{\ding{108}}), is affected by the added network delay.

Here, cloud-based deployment maintains adequate safety margins at lower speeds but begins to deteriorate at higher speeds. In one critical case involving a high-speed truck, the cloud platform produces a braking response that occurs within the unsafe zone. This outcome is not due to delayed detection ($\tau_{\text{det}}$), but rather a shortened actuation window ($\tau_{\text{react}}$) caused by tail latency, which proves insufficient for the longer vehicle deceleration profile. Interestingly, cloud deployment yields braking profiles that are quite similar to those of the on-device configuration, as seen by the overlapping brake reception distances. Further, we find that under extreme tail latencies ($p99$ and beyond), cloud deployment yields unsafe braking responses for heavier vehicles such as trucks even at moderate speeds ($40~\mathrm{mph}$). Although such latency spikes are rare, their occurrence poses a disproportionate risk for cloud-driven high-momentum scenarios.

\textbf{Takeaway: }These results underscore a key insight: early detection alone is insufficient, especially if control action is delayed. Therefore, platform selection for safety-critical applications must take into account both typical and worst-case network conditions. These empirical findings reinforce Lemma 2 and 3 by illustrating how delays introduced by communication can decouple perception from actuation, shifting the effective response time ($t_{\text{brake}}$) and potentially breaching safety margins. The effect is especially pronounced for vehicles with lower braking efficiency, where even small misalignments between detection and control can lead to unsafe outcomes.

\subsection{Impact of Concurrent Workloads}

The overall response time, ($T_{m,x}$), is most strongly influenced by the inference latency of the selected model-platform pair. Variations in $\tau_{\text{inf}}^{(m,x)}$ can arise in multitenant cloud environments, where resource sharing across diverse workloads leads to fluctuating GPU availability. While modern cloud systems employ load-balancing techniques to manage such variability, moderate levels of concurrent usage are still likely. For on-device deployments, we restrict our analysis to scenarios with minimal additional load. Even in dedicated hardware settings, some degree of parallel execution, such as concurrent sensing or lightweight control routines, may introduce non-negligible delays. This section evaluates how such concurrent workloads can influence downstream performance.

We begin by characterizing how inference latency and GPU utilization are affected by increasing concurrent workload on both cloud and on-device deployments. The measurements, shown in Figure \ref{fig:standalone}, were obtained using the \texttt{nsys} \cite{nsys} and \texttt{tegrastats} \cite{tegrastats} profiling tools, respectively, while varying the number of co-located YOLO11 inference servers from 0 to 10. As expected, the cloud GPU exhibits better scalability, handling additional clients with relatively modest increases in utilization and latency. In contrast, the on-device GPU demonstrates sharp latency spikes ($100-150~\mathrm{ms}$) even under low additional load. Although these trends are closely related to the hardware design of each platform, they provide a representative baseline for interpreting downstream differences in braking performance.

Under conditions of asymmetric load, that is, moderate concurrent inference load in the cloud and minimal load on the on-device platform, we observe nuanced trade-offs in braking performance. While the cloud configuration continues to demonstrate superior actuation margins at lower speeds, violations of the safety constraint emerge at higher speeds ($60~\mathrm{mph}$), where increased network and compute delays under load narrow the available reaction window, $\tau_{\text{react}}$. In contrast, the on-device platform, despite benefiting from zero communication latency and bounded inference delay, fails to meet the braking constraint at both moderate and high speeds due to its limited computational capacity. These results suggest that while cloud inference scales more gracefully with workload and offers tighter performance distributions at low to moderate speeds, it becomes vulnerable to safety violations under high-velocity scenarios where even modest latency increases can be detrimental. On-device inference, though more predictable, offers little margin for dynamic actuation under increasing speed or computational complexity. 

\textbf{Takeaway:} These findings corroborate Lemma~3 by demonstrating that temporal misalignment introduced by inference or network delay can compromise control safety, and neither deployment strategy is universally optimal. Instead, effective system design must consider the joint impact of model-platform characteristics, and operational and load conditions when selecting inference targets for real-time control tasks.

\subsection{Varying Environmental Context}

Until now, our analysis has focused on system-level factors that directly influence response latency and control performance. However, as formalized in Lemma 3, the environmental context plays a critical role in determining the timing of obstacle detection itself. We previously evaluated two such contextual variables: vehicle type, where differences in deceleration capacity affect stopping time ($t_{\text{stop}}$), and vehicle speed, which governs the margin available for safe braking ($\tau_{\text{react}}$). We now examine a third environmental factor: the size of the obstacle. This is closely linked to perception uncertainty, as smaller obstacles or those farther away occupy fewer pixels in the input frame, delaying confident detection. To study this, we vary the obstacle type by selecting a pedestrian (small), a bicycle (medium) and a car (large) as representative cases, and evaluate system performance at an ego vehicle speed of $40~\mathrm{mph}$, as shown in Figure 3(d). As expected, smaller obstacles are detected later (ie, have larger $t_{\text{det}}$), resulting in shorter stopping distances. Importantly, the cloud configuration is more resilient to such contextual variations, owing to its ability to deploy larger and more accurate models that not only complete inference faster, but are more sensitive to smaller changes in the scene. 

\textbf{Takeaway:} This experiment further illustrates how different environmental variables (vehicle type, speed, and obstacle size) modulate different components of the system's temporal pipeline, influencing when detection occurs and how much actuation time remains - both of which critically affect downstream control performance as expressed in Equation 1.
\section{Discussion and Conclusion}

Our work revisits the prevailing assumption that cloud-based inference is fundamentally too slow for real-time decision-making critical to safety. Rather than treating the cloud as inherently infeasible due to its physical remoteness and network delays, we offer a structured framework, grounded in analytical modeling and empirical validation, that identifies the conditions under which cloud inference is not only viable but preferable. We do not claim universal superiority of the cloud; instead, we articulate \textit{when} and \textit{why} it can outperform local execution in real-time control loops.

We formalize this investigation through a series of lemmas that characterize the effects of network latency, inference delay, and detection timing on control actuation. Each lemma is experimentally validated through controlled simulations that approximate real-world driving conditions. By testing across diverse speeds, vehicle types, and platform loads, we show that these analytical limits match the observed system behavior. Specifically, our results confirm that cloud inference can meet or exceed the safety performance of on-device systems, even for latency-sensitive applications like emergency braking, when model and queuing dynamics are properly accounted for.

Our modeling framework is deliberately conservative. It incorporates queuing delay on both cloud and on-device deployments, models detection delay as a function of model complexity and input sensitivity, and defines feasibility with respect to task-level physical constraints such as braking distance. These choices represent a departure from platform-centric narratives (e.g., "cloud is farther, therefore worse") toward task-aligned reasoning: Given the sensing frequency, the model execution profile and the actuation demands, can the system respond in time?

Empirically, we find that the cloud's ability to sustain higher service rates gives it a notable advantage under multi-tenant workloads. In such scenarios, even modest additional delays on the device can lead to constraint violations, while the elastic compute resources of the cloud allow it to maintain bounded inference delays. This scalability makes the cloud especially attractive for applications that must handle concurrency or operate under resource constraints.

Although our findings highlight the conditions under which cloud inference is feasible, they also expose scenarios where it falls short. A key failure case arises when stopping a heavy vehicle, such as a truck, at high speeds under high latency or workload conditions. These scenarios motivate hybrid architectures, where the cloud performs early, lower-accuracy detection and relays this information to the on-board platform for further processing. This cooperative approach can enable a timely braking response even under adverse conditions. Additionally, in safety-critical domains such as pedestrian detection, systems often rely on sensor fusion across modalities such as LiDAR and multi-camera arrays to improve robustness. Although our current analysis abstracts away this complexity, it remains representative of real-world deployments by capturing the timing and computational bottlenecks that ultimately govern feasibility.

Other system-level considerations that affect performance warrant further analysis. In particular, we assume a fixed frame rate $F$, yet real-world systems can dynamically adjust frame rate based on context, such as increasing it in high-speed scenarios. Such variability introduces additional queuing and tightens inference deadlines. Further, encryption of data in transit introduces additional overhead, which, while amortizable in persistent sessions, may further tighten feasibility bounds under high-frequency workloads. Moreover, dynamic obstacles introduce spatio-temporal uncertainty and require predictive models for motion planning and avoidance. These extensions would require integration of real-time tracking and trajectory forecasting modules and could impact the interplay between detection timing and control actuation. However, our formulation provides a foundational model that captures the dominant trade-offs in latency-sensitive perception and can be extended to accommodate these complexities in future work.

Despite these abstractions, the core insight remains: Cloud is not inherently disqualified by its distance. When inference delay, queuing behavior, and detection timing are modeled in a task-aligned fashion, cloud inference can meet, and sometimes exceed, real-time safety requirements. As network latencies decrease and model inference efficiency improves, the case for the cloud as a viable, even preferred, inference platform in safety-critical CPS becomes increasingly compelling.

Moreover, the insights developed here extend well beyond braking. Real-time CPS tasks such as adaptive cruise control, industrial automation, and medical intervention share common structural constraints: periodic sensing, latency-bound actuation, and feasibility thresholds. By integrating platform-level characteristics with application-level constraints, our framework offers a generalizable lens for reasoning about remote inference in these domains. By reframing viability as a function of context rather than location, this work lays the groundwork for more principled and scalable deployment strategies across a wide spectrum of real-time systems.

\section*{Acknowledgment}

This research was funded in part by DEVCOM ARL under the cooperative agreement W911NF1720196, and by the NSF under awards CNS-2211301 and CNS-2325956.

\bibliographystyle{ieeetr}
\bibliography{ref}

\end{document}